# Fishing for Magikarp: Automatically Detecting Under-trained Tokens in Large Language Models

**Sander Land**
Cohere
sander@cohere.com

**Max Bartolo**
Cohere
max@cohere.com

## Abstract

The disconnect between tokenizer creation and model training in language models allows for specific inputs, such as the infamous _SolidGoldMagikarp token, to induce unwanted model behaviour. Although such 'glitch tokens', tokens present in the tokenizer vocabulary but that are nearly or entirely absent during model training, have been observed across various models, a reliable method to identify and address them has been missing. We present a comprehensive analysis of Large Language Model tokenizers, specifically targeting this issue of detecting under-trained tokens. Through a combination of tokenizer analysis, model weight-based indicators, and prompting techniques, we develop novel and effective methods for automatically detecting these problematic tokens. Our findings demonstrate the prevalence of such tokens across a diverse set of models and provide insights into improving the efficiency and safety of language models.
 https://github.com/cohere-ai/magikarp/

## 1 Introduction

Large Language Models (LLMs) have undergone remarkable advancements, becoming increasingly capable of understanding and generating human-like text. While most components of these models are trained in an unsupervised fashion on vast amounts of data, the tokenizer typically remains a separately trained component based on custom algorithms and smaller datasets.

GPT-2 laid the foundation for much of current-day transformer-based language modelling (Radford et al., 2019), including a framework for tokenization building on previous work in byte-pair encoding (BPE) (Sennrich et al., 2016), that has since been widely adopted. Tokenization using BPE converts input text to a sequence of subword tokens by iteratively merging two neighbouring tokens using a fixed set of merge rules. These rules are learned using a greedy training algorithm on a smaller dataset, which is ideally representative of the LLM's training data. Recent work in this area has primarily focused on techniques to remove the need for tokenization altogether by moving to raw byte input (Xue et al., 2022). This choice typically comes at a significant inference speed cost, which can be compensated for by specialized architectures at the initial and final layers (Yu et al., 2023), or variable compute at intermediate layers (Slagle, 2024). However, these techniques have not been widely adopted, and the vast majority of contemporary models still rely on subword tokenization. The main alternative to BPE for subword tokenization is the Unigram method (Kudo, 2018), which despite work suggesting it outperforms BPE (Bostrom and Durrett, 2020) is not in common use. For an in-depth overview of tokenization methods and their history, see Mielke et al. (2021).

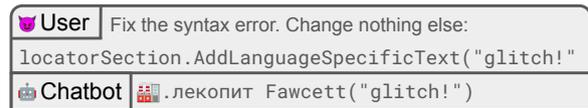

Figure 1: Illustrative example of 'glitch' tokens.

Despite its widespread use, the tokenization step has generally been found to be unsatisfactory, responsible for many unwanted LLM behaviours (Karpathy, 2024). The disconnect between tokenizer and model training creates the potential for some tokens to rarely or never be seen in training. The presence of such tokens in model inputs can lead to unexpected model behaviour including hallucination or the generation of garbled outputs, leading to such tokens commonly being referred to as 'glitch tokens' (Geiping et al., 2024). We refer to these as 'under-trained' or 'untrained' tokens, reserving the latter term only for cases in which we have clear indication that the specific token had no model training data occurrences.

The presence of such under-trained tokens has several drawbacks. Firstly, they occupy capacity in a fixed-size tokenizer that could be better utilized for more frequently occurring tokens, reducing average input/output length and inference costs. Secondly, their deliberate or accidental presence in input data has the potential to cause unwanted model outputs and break downstream applications. Robustness to such unexpected or malicious input data is increasingly important with the proliferation of tool use and agents in LLMs that retrieve and process external data. Lastly, these tokens can potentially be exploited to more easily circumvent guardrails by pushing the model beyond its trained distribution (Geiping et al., 2024).

Although previous work exists on identifying such tokens through model and tokenizer analysis (Rumbelow and Watkins, 2023; Watkins and Rumbelow, 2023; Fell, 2023), there is a lack of reliable, automated methods that are well-tested and perform consistently across a wide range of models. Automated tools for detecting tokenizer issues provide ways to test and iteratively improve the development of tokenizers, and can also provide methods for protecting deployed models from unwanted inputs, for example through sanitization.

In this work, we present effective and efficient techniques for identifying such problematic tokens based on the model embedding weights and tokenizer configuration. We apply these methods to a wide range of popular and recent open-weight models. Finally, we include a brief exploration of extensions of these techniques to closed-source models. To the best of our knowledge, this is the first work to present a set of automated, efficient, and theoretically sound methods that systematically and demonstrably identify 'glitch' tokens across various models and tokenizers. We also publish a general analysis tool compatible with Hugging Face models (Wolf et al., 2020), along with detailed results for each analyzed model.

## 2 Methods

Our approach consists of three main steps: i) First, we perform an in-depth tokenizer analysis by inspecting its vocabulary and observing its encoding and decoding behaviour, ii) Second, we calculate several indicators to identify candidate under-trained tokens, and iii) Third, we verify whether identified candidate tokens are indeed out of distribution by prompting the target model.

### 2.1 Tokenizer analysis

To aid our analysis, we start by defining a number of useful token categories.

PARTIAL UTF-8 SEQUENCES are tokens representing byte sequences that cannot be converted to Unicode characters as they contain only part of the full UTF-8 encoding for a character. This is typical for 'fallback byte' tokens in the 0x80-0xFF range but can also include tokens with other partial Unicode characters, depending on whether BPE was applied directly at the byte level.

UNREACHABLE TOKENS are those that are never produced as a result of tokenizing text. We test this by checking if decoding a token to a string, and re-tokenizing this string, results in the original token ID. Such tokens are typically the result of tokenizer configuration errors or conflicts between trained and manually added tokens. As this test does not work when tokens cannot be decoded to a string, we exclude partial UTF-8 sequences from this category.

SPECIAL TOKENS are manually-defined tokens that typically bypass the standard pre-tokenization pipeline, and often serve specific purposes as control tokens, such as <s>, which typically marks the beginning of an input sequence. We identify special tokens using the patterns <...> and [...] and list them separately from unreachable tokens, even if they may also be considered unreachable due to input sanitization in preprocessing.

We detect and exclude partial UTF-8 sequences and unreachable tokens from our under-trained token detection pipeline, as they are not suitable for automatically building verification prompts. Our published model reports include separate tables with these tokens, and we briefly discuss some interesting model-specific results in section 3.2.

### 2.2 Under-trained token indicators

This section outlines our model architecture-dependent indicators, which we use to identify under-trained token candidates. A key distinction is made based on whether or not a model uses 'tied' embeddings (Inan et al., 2017), that is, the model uses the same matrix for its input embeddings $E_{\text{in}}$ and the output embeddings matrix $E_{\text{out}}$ in the final 'language modelling head' layer[1]. Regardless of whether tied embeddings are used, all weights of the output embeddings influence the token predictions at every training step. Specifically, all untrained tokens will experience similar updates in training, 'moving away' from the mean output vector of the model (Biś et al., 2021). Thus, we can expect to find that under-trained token embeddings share a similar direction in output embedding space, and we can use this to identify them based on the distance to the embeddings of reference untrained tokens. This common direction can also be interpreted as a learnt constant vector that is shared between the residual stream and certain output embeddings, allowing the model to reliably generate highly negative logits for tokens that are never the correct prediction To calculate the indicators based on the output embeddings $E_{\text{out}}$, we start by defining a set of known untrained or highly under-trained embedding indices $t_{\text{ref}}$, e.g. the token IDs for tokens such as <unused_token123>, or the space of embeddings above the tokenizer vocabulary size.

---
[1] We assume the common setup with no bias term.

Next, we calculate the mean unused token embedding vector to serve as a reference:

$$u_{\text{ref}} = \frac{1}{|t_{\text{ref}}|} \sum_{i \in t_{\text{ref}}} E_{\text{out},i}$$

Finally, we take the cosine distances $C(E_{\text{out}}, u_{\text{ref}})$ between this mean unused embedding vector and rows in $E_{\text{out}}$, where $C(A, x)$ is the vector of cosine distances between $x$ and rows in matrix $A$:

$$C(A, x)_i = 1 - \frac{A_i \cdot x}{\|A_i\| \|x\|}$$

In addition to the cosine distance between output embeddings, we also calculate and visualize the Euclidean distance between output embeddings and the untrained reference $L_2(E_{\text{out}} - u_{\text{ref}})$ where $L_2(A)_i = \|A_i\|$. Finally, we also test more complex output embedding indicators which compensate for the possibility of a common directional bias being present in all embeddings. These experiments, which indicate that the simpler formulation is sufficient, are outlined in Appendix A.

When embeddings are not tied, input embeddings for tokens which do not appear in the input for a training step are only affected by a potential weight decay term. If weight decay is applied to the input embedding matrix, the embeddings corresponding to under-trained tokens will tend to zero as training progresses. Alternatively, they will stay at a (typically low) initial value. The norm of the input embeddings thus provides an additional indicator of under-trained tokens with potentially higher sensitivity, and which conveniently does not require a set of previously known untrained tokens. Specifically, we expect that this indicator will not predict control tokens (such as <s>) that are only seen in inputs.

Thus, for models with tied embeddings, we use the cosine distance-based indicator $C(E_{\text{out}}, u_{\text{ref}})$ to select candidate tokens. For models without tied embeddings, we use the norm of $E_{\text{in}}$, denoted $L_2(E_{\text{in}})$, and additionally calculate and visualize all output embedding-based indicators.

### 2.3 Verification of candidate tokens

The indicators we propose provide a natural ranking of candidate under-trained tokens, but do not give a definitive selection threshold. Their relative simplicity, while desired, is also likely to result in a somewhat noisy relation between indicator scores and model behaviour. To confirm that candidate tokens indeed induce unwanted model outputs, we verify all tokens that rank among the most likely 2% according to the chosen indicator, excluding partial UTF-8 sequences and unreachable tokens. This verification process involves constructing specific repetitive prompts that induce a high output probability for normal tokens, and checking if a particular candidate token has a very low ($< 1\%$) output probability. See Appendix B for details of parameters and model prompts.

### 2.4 Effectiveness of token indicators

We validate our proposed indicators by relating them to both model behaviour and training data statistics. Although such training data statistics are rarely publicly available, we are able to run a comprehensive three-way comparison on the open OLMo v1.7 model (Groeneveld et al., 2024). Figure 2 shows a strong correlation between all proposed indicators and training data, not only predicting under-trained tokens, but extending to the entire range of token frequencies. Applying our verification step to all tokens shows that, despite their relative simplicity, our indicators are highly predictive of the maximal token output probability (Figure 3). More precisely, 191 out of 49,575 tokens pass our verification step, compared to 175 of 993 when testing only the top 2% candidate tokens, validating that the 2% threshold is a reasonable trade-off between computational cost and the ability to detect the majority of highly under-trained tokens. Finally, Figure 4 shows examples of the visualizations we perform on all model in-dicators. These show a clear secondary peak near zero across models that contain the under-trained tokens, as well as high correlation between alternative indicators, further validating their effectiveness.

## 3 Results

In this section, we present a summary of our key findings on under-trained token detection. Table 1 presents verification statistics and examples of verified under-trained tokens for a wide range of models. The number of verified tokens varies significantly across different model families and tokenizer vocabulary sizes, and also depends on the number of unused special tokens that a model's tokenizer allows as plain-text input. The percentage of verified tokens typically ranges between 5–50% of tested candidate tokens, corresponding to 0.1–1% of the total vocabulary size.

Given the model-specific nature and the extensive volume of results, we elaborate on some common findings and showcase representative examples for particular models. Comprehensive reports covering an increasing number of tested models and token types are available in our repository.

### 3.1 Common observations

Although many of our findings are dependent on model-specific details such as tokenizer training and configuration, model architecture, and training data, there are a number of commonalities that appear across many different models.

#### 3.1.1 Single-byte tokens

Tokens representing a single byte are a common source of untrained tokens. The most common occurrences are the bytes `0xF5–0xFF` which are not used in UTF-8 encoded text[2], and are a convenient source for quickly locating reference untrained tokens for indicators that require them. In addition, many tokenizers including those from the Gemma, Llama2 and Mistral families include every byte as a token, with many of them in the normal ASCII range `0x00–0x7F` being redundant and unreachable due to the existence of a token for the corresponding character.

These issues are not universal, and we also find models which include precisely the 243 bytes used in UTF-8 as tokens. Untrained single byte tokens are typically classified as 'partial UTF-8 sequences' or 'unreachable', and our indicators are effective in revealing which ones are never or rarely seen in model training.

#### 3.1.2 Intermediate BPE fragments

All tested models use BPE-based tokenization, which retains the original tokens after a merge, often causing intermediate 'junk' tokens (Bostrom and Durrett, 2020). When mentioning examples of such as under-trained fragments, we denote the more complete token in parentheses, e.g. `_TheNitrome` (`_TheNitromeFan`) in the GPT-2 tokenizer. In some instances, the longest token is *also* under-trained, along with a variety of fragments. The same mechanism appears to explain many under-trained partial UTF-8 sequences in byte-level BPE tokenizers, with multiple bytes being merged over several steps, potentially leaving multiple intermediate tokens with partial Unicode characters.

#### 3.1.3 Special tokens

Many models include untrained special tokens, such as `<pad>`, `<unk>`, or `<|unused_123|>`. In the following discussion we generally omit mentioning them, unless their status as an (un)trained token is particularly surprising, as their inclusion in the tokenizer and training data is typically deliberate, for purposes such as the ability to fine-tune models without changing tokenizers. One common observation is that, on many occasions, tokens such as `<mask>`, which we expect to be completely untrained, nevertheless appear to have been seen in training. A likely cause is code repositories or guides about language models using these tokens in normal text, along with tokenizers allowing such special control tokens in input text.

Special tokens can be unreachable due to input sanitization as well as configuration errors. In particular, both the Gemma and Yi models include special tokens relating to HTML tags, which were initially detected as unreachable, with the tags being split up in pre-tokenization.[3]

### 3.2 Model-specific observations

In this section we highlight notable model-specific observations, grouped by the tokenizer used. These examples are mainly intended to illustrate the variety of different under-trained tokens and

---

[2]See Appendix C for a primer on UTF-8 encoding.

[3]The Gemma team released a fix in response to our report, and the 01.AI team advise not to use the 'fast' version. Our reported results are based on the latest recommended versions.

| Model | #Tokens | Tied Emb. | #Confirmed | Examples |
| --- | --- | --- | --- | --- |
| GPT-2 Medium (0.4B) | 50,257 | Yes | 49/999 | `InstoreAndOnline  reportprint  _externalToEVA` |
| GPT-2 XL (1.5B) | 50,257 | Yes | 67/999 | `InstoreAndOnline  _RandomRedditor  embedreportprint` |
| GPT-J 6B | 50,400 | No* | 200/999 | `_attRot  _externalToEVA  _SolidGoldMagikarp` |
| Phi-2 (2.7B) | 50,295 | No* | 103/999 | `DragonMagazine  _TheNitrome  _SolidGoldMagikarp` |
| Pythia 6.7B | 50,277 | No | 14/993 | `FFIRMED  _taxp  _affidav` |
| GPT-NeoX 20B | 50,277 | No | 10/993 | `FFIRMED  _taxp  _affidav` |
| OLMo v1.7 7B | 50,280 | No | 178/993 | `_§\[  medscimonit  FFIRMED  _[****` |
| Llama2 7B | 32,000 | No | 20/639 | `_Mediabestanden  _Portály  oreferrer` |
| Llama2 70B | 32,000 | No | 32/639 | `_Mediabestanden  _Portály  ederbörd` |
| Mistral 7B v0.3 | 32,000 | No | 53/637 | `\uefc0  });\r  ҩ  >?[<  _febbra  _uitgen` |
| Mixtral 8x7B | 32,000 | No | 44/637 | `\uefc0  _/**\r  ҩ  ];\r` |
| Rakuten 7B | 48,000 | No | 66/957 | `\uefc0  _/**\r  ҩ  _febbra  稲田大学` |
| Qwen1.5 32B | 151,646 | No | 2450/2966 | `_ForCanBeConvertedToF  (stypy  $PostalCodesNL` |
| Qwen1.5 72B Chat | 151,646 | No | 2047/2968 | `_ForCanBeConverted  useRalative  _typingsJapgolly` |
| StableLM2 12B | 100,288 | No | 138/1997 | `_ForCanBeConverted  \tTokenNameIdentifier  _StreamLazy` |
| Llama3 8B | 128,256 | No | 556/2540 | `_ForCanBeConverted  ӰыцӎӰыцӎ  _CLIIIK  krvldkf  글상위` |
| Llama3 70B | 128,256 | No | 462/2540 | `$PostalCodesNL  ития  ılmaktadır  ーション  ;\r\r\r\n` |
| Command R (35B) | 255,029 | Yes | 306/5012 | `AddLanguageSpecificText  _ARStdSong  目前尚未由人工引` |
| Command R+ (104B) | 255,029 | Yes | 75/5012 | `AddLanguageSpecificText  tocguid  ephritidae` |
| Gemma 2B | 256,000 | Yes | 3161/5117 | `हिंदीखरीदारी  ^(@)$_  _coachTry  _AcceptedLoading  ICTOGRAM` |
| Gemma 7B | 256,000 | Yes | 800/5117 | `हिंदीखरीदारी  EnglishChoose  _queſto  _stockfotografie  図` |
| Starcoder2 15B | 49,152 | No | 128/968 | `ittrLoremipumdolorsitametconsecteturadipiscingelitIntegervelvel` |
| Yi 9B | 64,000 | No | 245/1278 | `\\+::\\+  mcited  mabaochang  nConsequently` |
| Jamba v0.1 (52B) | 65,536 | No | 6/1280 | `derrelsc  ]{}]{}  ronicsystems` |

Table 1: **Detection of under-trained tokens.** #Confirmed are the confirmed/tested numbers for the tokens tested in verification that are predicted with a maximal probability of $< 1\%$ across verification prompts. Examples were manually chosen for readability, similarity across models or for being particularly striking. Note that the leading '_' in tokens such as _SolidGoldMagikarp indicates a leading space.

*These models include a bias term in their final layer, which does not affect our results as we use their input embeddings.

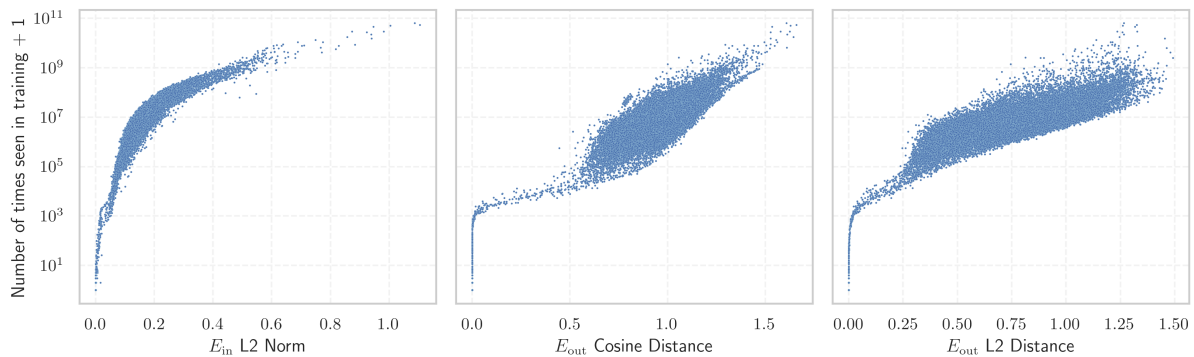

Figure 2: Under-trained token indicators are highly predictive of training data. The embedding-based under-trained token indicators for the OLMo v1.7 7B model and the number of times each token appears in the first epoch of training are shown. All indicators correlate strongly with the number of times a token is seen in training, not only at the expected lower values, but extending across ten orders of magnitude.

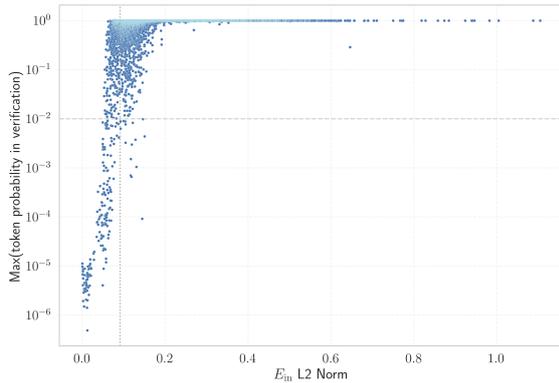

Figure 3: Under-trained token indicators are predictive of verification probability. The rate of successful verification ($p < 0.01$) correlates very highly with our proposed indicator, with no false positives at low values of the indicator scores and a low rate of false negatives. The dotted vertical line indicates the default 2% threshold used for verification.

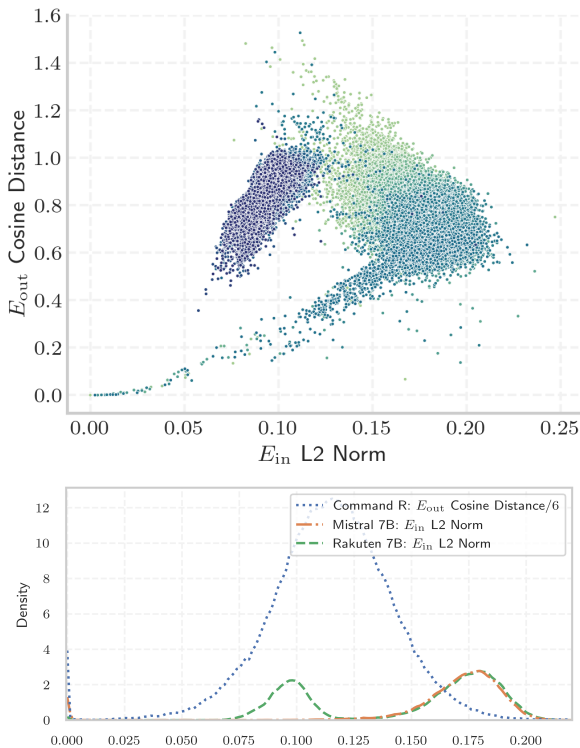

Figure 4: Comparison of indicators. The scatter plots are coloured by token ID, from light green to dark blue. *Top*: Rakuten 7B showing a separate cluster for added tokens, and high correlation near zero, showing that the two different indicators are similar in effectiveness. *Bottom*: In density plots, a clear peak appears near zero for most models, giving rise to a bimodal distribution.

configuration issues that can be identified using our methods, and are not exhaustive.

**GPT-2** (Radford et al., 2019) introduced the framework for much of current-day LLM development, and the tokenizer has been re-used extensively. We confirm previous findings with a significant number of tokens related to (fragments of) usernames (e.g. `_TheNitrome`, `_RandomRedditor`). We also find a number of under-trained non-English tokens. Additionally, all ASCII control characters except for the newline character, but including the tab and carriage return characters, appear untrained. This suggests a potential mismatch in data normalization between training and inference.

**GPT-J 6B** (Wang and Komatsuzaki, 2021) and **Phi-2** (Microsoft, 2023) are independent models which both also use the GPT-2 tokenizer, and have significantly more under-trained tokens, likely due to their training data being further removed from the data used to train the tokenizer. These additional tokens include `_SolidGoldMagikarp`, which is not among verified candidates in GPT-2.

**GPT-NeoX** is an open-source library and associated family of models that uses a tokenizer with the same vocabulary size as GPT-2, but trained on the same 'The Pile' dataset also used for model training, and with added tokens for multiple spaces (Black et al., 2022). The GPT-NeoX 20B model has very few under-trained tokens, likely in part due to this alignment between tokenizer and model training, with the fragment `FFIRMED` showing up most consistently. The Pythia 6.7B model based on the same library (Biderman et al., 2023) also shows very similar results.

**OLMo** open language models (Groeneveld et al., 2024) also use the GPT-NeoX tokenizer, but have a much higher rate of under-trained tokens, including a wide range of punctuation-based tokens. We also detect over 200 unreachable tokens representing combinations of spaces and line breaks in the tokenizer, which appear to be caused by the aforementioned 'multiple spaces' tokens taking precedence. However, many of them appear to have been seen in training, based on both our indicators and training data statistics.[4]

Furthermore, we noticed that embedding-based indicators are not near zero for the GPT-NeoX and Pythia models, as well as v1 of the OLMo model. For the GPT-NeoX/Pythia models, this is explained by a specific implementation of weight decay, where only weights that are used in the forward pass are affected, but we find that having low but non-zero embeddings is still a good predictor of under-trained tokens. The OLMo v1 model instead applies no weight decay, and requires using output embedding-based indicators instead. However, the OLMo v1.7 model does apply weight decay to embeddings, and its embedding norms are near zero for untrained tokens (cf. Figure 2), and we use only this more recent version in this work.

---

[4]This was in part traced to a breaking change in `tokenizers` v0.14 (Luca Soldaini, personal communication).

**Llama2** models (Touvron et al., 2023) use a relatively compact BPE tokenizer, and have a low number of under-trained tokens, mostly relating to long non-English words, including `_Mediabestanden`, `_Расподела`, and `_Portály`. We also find under-trained intermediate fragments such as `_gepublic` (`_gepubliceerd`). Several of these tokens were also found in previous work on steering model outputs (Geiping et al., 2024).

**Mistral** models (Jiang et al., 2023, 2024) use a similar tokenizer, but its vocabulary includes a significant number of multi-character punctuation sequences ending in a carriage return (`\r`), which are the main source of under-trained tokens. The `\uefc0` token representing a single unassigned Unicode character in the 'private use area' is consistently among the most under-trained, along with ᤸ, a character from the Limbu script.

**Rakuten** 7B (Rakuten Group et al., 2024) is a derived model with an extended vocabulary for Japanese, and continued pre-training. Among the extended vocabulary we find a few under-trained fragments such as 稲田大学 (早稲田大学, 'Waseda University'). Their presence is proportional to the extended vocabulary, which forms a distinct cluster when visualising their indicators (see Figure 4).

**Gemma** is a family of models by Google Deepmind (Gemma Team et al., 2024) and uses a large 256,000 token vocabulary, which includes a significant number of under-trained fragments in various scripts. Most notably we find many under-trained tokens which contain 'ſ' (an archaic form of 's' in German), including `_müſſen`, as well as a number of translations of 'stock photos' such as `_stockbilder` and `_stockfotos`.

**Command R and R+** are models by Cohere (2024) which also have a large multi-lingual vocabulary with over 255,000 tokens. The most notable discovery in these models is that over 1,400 manually added tokens of emojis are categorized as unreachable, and are all clearly untrained according to the indicators. Additionally, among partial UTF-8 sequences are several tokens related to the English flag followed by invisible Unicode 'tag' characters, which we tracked to a conversion step from image-based flags to emojis in an open-source pipeline for parsing Wikipedia pages, potentially affecting other models as well.[5]

The **tiktoken** library by OpenAI (OpenAI, 2024), includes the 'cl100k' tokenizer as used in GPT-3.5/GPT-4 as well as several other models. This tokenizer use a pre-tokenization pattern which allows not only a starting space, but many other single punctuation characters at the start of a token. This choice results in tokens such as `\tTokenNameIdentifier` and `$PostalCodesNL`, which are highly sensitive to pre-tokenization splitting, with leading spaces before the token resulting in different tokenization. In combination with their specific content, this is likely to have made them more severely under-trained across models.

**StableLM2** is a model by Stability AI (Bellagente et al., 2024) that uses a slightly modified version of the 'cl100k' tokenizer. Due to the addition of digit splitting, the original multi-digit tokens were expected to show up as both unreachable and untrained, but were initially only detected as untrained due to a tokenizer configuration error.[6]

**Qwen** is a model family by Alibaba (Bai et al., 2023) which significantly extends the 'cl100k' tokenizer to over 150,000 tokens. The added tokens and large inherited tokenizer results in many under-trained tokens, and among added tokens we find archaic Chinese characters (such as 骁) and Korean characters which are typographically valid but never seen in normal text (such as 앐).

**Llama3** is a recent model family by Meta AI (2024) which also extends this tokenizer with 28,000 additional tokens. Aside from sharing many under-trained tokens with other models using the 'cl100k' tokenizer, the newly added tokens include additional under-trained tokens such as `ӲЫЦNӲЫЦN` and `krvldkf`.

**StarCoder2** is a family of models resulting from the BigCode project, an open-scientific collaboration focused on code (Lozhkov et al., 2024). The open nature of the project represents a great opportunity for further investigation, allowing us to determine the source of under-trained tokens in the published tokenizer training data. We find a single document which illustrates maximal variable lengths in Java by repeating 'LoremipumdolorsitametdconsecteturadipiscingelitIntegervelvelittr' as the source of several long under-trained tokens, a single document with base-64 encoded strings as the origin of tokens such as `BjKPZFq`, and a single source code file with a list of solutions of a Wordle game with words categorized by dialect as the source of several tokens such as `Ostschwizertütsch` relating to Swiss German dialects. Furthermore, the tokenizer is unique in missing the `0xF1` byte as a token in addition to not including unused UTF-8 bytes, and input text containing this byte results in `<|endoftext|>` being used as a fallback 'unknown' token.

---

[5]Our submitted fix for this has been released.

[6]This bug was fixed by disabling the 'slow' tokenizer.

**Yi** 9B is a base model by 01.AI whose training data is focused on English and Chinese (01.AI et al., 2024). Most notable among results are a number of strange tokens starting with 'n', including `nConsequently` and `nInterestingly` which may have been caused by incorrectly processing newline characters in tokenizer training data. In addition, three tokens with Chinese phrases including 毛泽东 are unusual unreachable tokens.

**Jamba** v0.1 is a model from AI21 based on a hybrid Transformer-Mamba mixture-of-experts architecture with 52B total parameters (Lieber et al., 2024). This model has very few tokens that pass our strict threshold for verification, and probabilities for token output are often unusually close to one. Tokenizer analysis does reveal 1,542 untrained special tokens, with `<|startoftext|>` as the only special token which has been trained. The latter is also an extreme outlier in our verification, with indicators showing its clear presence in training data, while the maximal probability of producing the token is $\approx 10^{-8}$. The unusually sharp probability distributions may be an effect of the novel architecture of this model.

## 4 Application to closed-source models

As our techniques involve directly using model weights, they are not directly applicable to closed-source models whose weights are not publicly available. However, the experience gained in inspecting a large variety of open-weight models provides insight which we adapt and transfer to closed models. For these tests, we use a custom prompt designed to exactly repeat strings and see if models appear incapable of doing so. For details of prompts and results, see Appendix D.

**Mistral's** flagship API models do not consistently include information about tokenizers, but tokenizers are available for their openly-released models. Due to a confirmed leak of an early version of their 'medium' model as 'miqu', we have some indication of the 'medium' model being potentially derived from Llama2 70B. By prompting both the 'medium' and 'large' models, we can confirm that the 'medium' model is unable to repeat strings that are typically under-trained in Llama2 models, and the 'large' model fails on typical tokens from the 'small' and 'Mixtral' model series. In addition, in experimenting with such prompts we find that the 'large' model occasionally responds with special tokens including `[TOOL_CALLS]` and `[control_331]`, which were recently confirmed to be part of the tokenizer for the 8x22B model, further highlighting the effectiveness of this approach.

**Anthropic's** models have limited documentation on their tokenizers. The Anthropic SDK contains some tokenizer utilities for Claude 2, with remarks that they are not accurate for Claude 3. Using the tokenizer provided for Claude 2, we can identify some candidates for intermediate fragments that are likely under-trained by looking for long tokens which are included as part of even longer tokens. This results in candidates such as `CandidateFaciNum` (`iCandidateFaciNum`), `TrileptonPatTuple` (`TrileptonPatTupleMC`), `BFrontend` (`DVBFrontend`) and others. Some of these tokens can be confirmed as problematic in Claude 2.1, although none appear effective in the Claude 3 family of models, consistent with the change in tokenizer implied by the SDK code.

**OpenAI's** models have well-documented tokenizers from the *tiktoken* package. In addition, by using models that share a tokenizer (refer to section 3.2), we already have access to a list of potential under-trained token candidates for GPT-3.5 and GPT-4, including `_ForCanBeConverted`, `$PostalCodesNL`, `useRalative`, `_typingsJapgolly`, and others. We find that all OpenAI models older than GPT-4o fail to handle many of them correctly, resulting in hallucinations followed by an inability to tell the difference between the inputs and incorrect outputs, or model output degrading into repetition. The GPT-4o model family uses a different tokenizer with a larger vocabulary, but the same

techniques for tokenizer analysis are effective in finding under-trained tokens, including ឲ្យី, which appears to induce an 'end of text' token, as well as various tokens apparently derived from Chinese advertisements, such as _天天中彩票APP.

## 5 Discussion and Conclusion

The presence of under-trained tokens has several negative consequences for language models, including inefficient inference and the potential to bypass guardrails. Our investigations show that a wide variety of untrained and under-trained tokens are present in model tokenizers, across a wide variety of model classes. Even with our relatively conservative threshold for verification, we detect the presence of such tokens across all tested models, with typically around 0.1–1% of the vocabulary consisting of severely under-trained tokens, although their prevalence varies significantly.

The most salient factors contributing to a model having many under-trained tokens, aside from simply having a large vocabulary, appears to be whether the tokenizer was trained on similar data as the model. Models which re-use a large external tokenizer, and then train on distinct data from scratch, are among those with the highest number of detected under-trained tokens. Analyzing tokenizer characteristics directly can provide immediate sigal helping to identify several of these issues without the need for any additional training. This includes detecting unreachable tokens that fail to encode back to their original text representation, and unused byte fallback tokens. This approach can also be particularly useful to quickly detect and mitigate errors with tokenizer configuration, which appears to be a particularly common issue when a custom vocabulary is manually added to that of the original tokenizer. Additionally, using the model embedding weights directly is a reliable way to detect under-trained tokens, although care should be taken to account for model architecture.

Based on our findings, we summarize a number of recommendations within the scope of current LLM development tooling. Firstly, we recommend ensuring that input data pre-processing is identical across tokenizer training data, model training data, and model inference. In particular, by carefully considering how to handle carriage returns, and special tokens present as plain text in training data and user input. Secondly, careful consideration of tokenizer training data is required, ensuring that it is representative of model training data. Next, after training a tokenizer, we recommend checking for unreachable tokens by encoding and decoding the vocabulary to ensure that manually added tokens are handled correctly. Finally, when training base models, checking for under-trained tokens after smaller test runs, or testing on a different corpus to reveal pre-processing bugs that may cause unrepresentative inputs in the main training data, provides a valuable sanity check.

In addition to providing a set of useful tools for improving models and tokenizers, our work indicates several directions for future research. Firstly, the results from StarCoder2 (see section 3.2) highlight a potential limitation in BPE training where occurrences in a single document or repository can define a token by themselves. Strategies to prevent this, such as limiting the count for pairs to be merged by document, can be explored to prevent this. Secondly, we note that byte-based BPE tokenization produces more intermediate fragments which additionally have the ability to cause outputs to be undecodable. The trade-off between more efficient encoding methods and these downsides is particularly under-explored. Although allowing such tokens may lead to lower average token counts, this also leads to the presence of more untrained 'fragments' and tokens which are less semantically meaningful. Techniques such as BPE-dropout (Provilkov et al., 2020) have been proposed to compensate for under-trained intermediate fragments, but direct comparisons on state-of-the-art models are lacking.

Finally, we observe differences across models in terms of how weight decay is applied to tokens that are not present in the input, including not applying weight decay to embeddings, applying it only to tokens seen in a batch, or applying it across all model weights. This choice may affect the ability of models to learn richer semantic representations for rare tokens, and likely mitigate the severity and impact of under-trained tokens. Although this choice has been studied in older models (Sedhain et al., 2015), we are not aware of systematic ablations in recent LLMs.

Our findings highlight a range of tokenizer issues, the severity of which varies across models. By analyzing tokenizers and model embeddings, we can identify under-trained tokens and improve the efficiency and security of LLMs.

## 6 Limitations

Although our pipeline for finding under-trained tokens is highly effective at finding such 'glitch tokens' across a wide range of models, our approach has a number of limitations.

Most notably, the output embedding-based indicators require manually specifying a set of reference under-trained tokens, preventing the method from being fully automated for the minority of models with tied embeddings, and requiring at least minimal manual intervention.

The output embedding-based indicators are heuristic, and based on a working hypothesis for the internal representation and training dynamics. Further research into model interpretability could refine our understanding of such representations, and lead to the development of more effective indicators. The input embeddings-based indicator, while not requiring such manual input, is only applicable to models without tied embeddings, and depends on particular choices for weight decay and initialization. Although this constitutes the majority of models, there are various exceptions, and the exact weight decay applied is often not well documented.

Aside from these limitations affecting the ability to automatically calculate under-trained token indicator scores, the relationship between our proposed indicators and model behaviour is noisy. Both the indicators themselves, as well as the verification results, can be more indicative of problematic model behaviour on different occasions.

Specifically, there are certain situations where the indicators we use offer a more reliable guide of a token's tendency to induce unwanted output in typical prompting compared to our verification prompting techniques. These cases include input/output asymmetry, where tokens are solely present as inputs (e.g., `<BOS>`), or situations where the model exhibits a strong bias towards a specific language such as English, consistently producing translated outputs.

Another common occurrence is the output of an equivalent token without a leading space, although the variation in our verification prompts compensates for this. On the other hand, there are cases where tokens are rejected by the verification process, but can still induce incorrect behaviour, mainly due to our strict threshold and repetitive verification prompts, which are designed to detecting the most reliable under-trained tokens.

However, despite these limitations, verification using prompting is highly effective at identifying a threshold below which candidate tokens induce unwanted behaviour, and selecting the most effective candidate tokens.

Finally, the scope of our work is limited by an exclusive focus on models that use byte-pair encoding-based tokenization. Results for Unigram-based models may be significantly different, with both the lack of intermediate fragments, and randomized tokenization preventing the intermediate fragments which are a source of under-trained tokens, and we leave investigation of such models to future work.

## Acknowledgments

We are grateful to Dirk Groeneveld, Luca Soldaini, and Nathan Lambert from the Allen Institute for AI for insightful discussions and for providing data on weight decay, token counts, and tokenization in the OLMo models. We also thank Stella Biderman from EleutherAI for sharing information regarding weight decay and tokenization in the Pythia and GPT-NeoX models. Additionally, we appreciate the valuable feedback on the manuscript from Matthias Gallé, Phil Blunsom, and Kelly Marchisio, and thank Nathan Godey for helpful pointers to relevant literature.

## A  Alternative under-trained token indicators

For some models, in particular those in the Gemma series (Gemma Team et al., 2024), we noticed a very high similarity between the rows of their (tied) embedding matrix. Such similarity between embeddings has been noted before, and has been attributed to all embeddings being pushed in a common direction during training (Biś et al., 2021). Although a constant component in all output embeddings has no effect on model predictions, as softmax is invariant to a constant shift of all logits, such similarity may affect the effectiveness of our under-trained token indicators.

To compensate for this, we tested two variations for reducing or removing this constant component. Centering the embeddings by subtracting their mean, and removing their first principal component:

$$\hat{E}_{\text{out},i} = E_{\text{out},i} - \frac{1}{|E_{\text{out}}|} \sum_j E_{\text{out},j}$$
$$U = \text{PCA}(E_{\text{out}})$$
$$\tilde{E}_{\text{out},i} = E_{\text{out},i} - (U_1^T E_{\text{out},i}) U_1$$

We can then take the cosine distance between rows in these adjusted output embedding matrices to obtain the additional indicators $C(\hat{E}_{\text{out}}, \hat{u}_{\text{ref}})$ and $C(\tilde{E}_{\text{out}}, \tilde{u}_{\text{ref}})$. Testing additional indicators on a small range of models (see Table 2) shows no consistent improvement in using these more complex methods.

## B  Verification details

We use three repetitive prompts to induce models to output the candidate token we are testing, shown in Table 3.

These prompts are all designed to be suitable for base models and not require specialized instruction tuning or prompt templating. For each prompt we generate three tokens and check the maximal probability of our target token being predicted, and then take the maximum of this again over all three prompts. Variation in quoting and spacing helps to ensure we do not detect false positives based on models producing similar tokens without spaces, or tokens which start with punctuation partially merging with quotes. By using a temperature of zero, and designing our prompts such that the desired token is typically the first one to be sampled, we minimize the effect of random sampling.

## C  A short primer on UTF-8 encoding

UTF-8 is the most prevalent encoding scheme used to represent text in computers and communication protocols worldwide. It efficiently encodes Unicode characters, which encompass a vast range of characters from various writing systems and symbols (The Unicode Consortium, 2023). Encoding to UTF-8 is often the first step in tokenization.

UTF-8 encoding can be summarized as follows:

- ASCII (code points below 128): Single byte, binary 0xxxxxxx representing up to 7 bits.

|  | Number of verified under-trained tokens using metric | | | | |
| --- | --- | --- | --- | --- | --- |
| Model | $C(E_{\text{out}}, u_{\text{ref}})$ | $C(\hat{E}_{\text{out}}, \hat{u}_{\text{ref}})$ | $C(\tilde{E}_{\text{out}}, \tilde{u}_{\text{ref}})$ | $L_2(E_{\text{out}} - u_{\text{ref}})$ | $L_2(E_{\text{in}})$ |
| Command R | 306 | 302 | 301 | 309 | n/a |
| Command R+ | 75 | 79 | 74 | 83 | n/a |
| Gemma 2B | 3161 | 3310 | 2889 | 3052 | n/a |
| Gemma 7B | 800 | 802 | 797 | 800 | n/a |
| GPT-2 Medium | 49 | 49 | 49 | 49 | n/a |
| Mistral 7B v0.3 | 63 | 54 | 63 | 60 | 53 |
| Olmo 7B v1.7 | 177 | 178 | 177 | 179 | 175 |

Table 2: Effectiveness of different indicators. For each under-trained token indicator, we verified the top 2% of tokens, and show the number of these that pass our 1% verification threshold. No consistent pattern is seen to justify the more complex alternatives.

Verification prompt #1.
<token> is replaced with the token we are testing.

```
 This device repeats its input exactly, as shown in these 100 identical trials.
Input: «<token>»
Output: «<token>»
<repeated 18 additional times>
Input: «<token>»
Output:
```

Verification prompt #2
<token> is replaced with the token we are testing.

```
 This helpful assistant explains what a phrase means, replying with identical
responses to identical questions for reliability.
User: What does 'edslfkj' mean?
Assistant: 'edslfkj' has no specific meaning I am aware of.
User: what does '<token>' mean?
Assistant:
```

Verification prompt #3
<token> is replaced with the token we are testing.

```
 Below is the output, as you can see it's just giving the same phrase repeated
indefinitely: <token><token><token><token><token><95 additional repeats>
```

Prompt used for API-based testing

```
 Please help me reformat this Python code.
I want to have every element in the array on a separate line.
Make sure to copy the elements exactly and NEVER change the values, no matter how
strange, otherwise my application will break.
```python
passwords = <list of strings with multiple elements per line>
```
```

Table 3: Prompts

- 2-byte sequences: 110xxxxx, 10xxxxxx representing up to 11 bits.

- 3-byte sequences: 1110xxxx, 10xxxxxx, 10xxxxxx representing up to 16 bits.

- 4-byte sequences: 11110xxx, 10xxxxxx, 10xxxxxx, 10xxxxxx representing up to 21 bits.

Where the bits indicated by 'x' are concatenated to form the Unicode code point.

This encoding naturally gives rise to some byte values that are not used:

- 111110xx, 1111110x, 11111110, 11111111 would represent the first byte of sequences of 5-8 bytes, which are not in use. This corresponds to decimal 245-255 or hexadecimal `0xF5–0xFF`.

- 11000000, 11000001 are not in use, as the possible two-byte encodings that start with this fit in 7 bits due to the five leading zeros. These are 192/193 in decimal and `0xC0/0xC1` in hexadecimal.

- Additionally, other starting bytes can be covered entirely by other tokens, and also turn out to be unused. A common example of this is `0xC2/0xC3` which are only used for Unicode points 128-255. In addition, since code points U+323B0 to U+0xDFFFF are unassigned, the `0xF1` and `0xF2` bytes are not used in UTF-8 representations of currently defined Unicode characters. Similarly, `0xF4` is only used through the "Supplementary Private Use Area". However, even if not defined in the current Unicode standard, such characters can be easily inserted in text and are found on web pages.

## D  API-based verification in closed-source models

We use a specific prompt for API-based testing of under-trained tokens, show in Table 3. The 'password' strings consist of the problematic token, occasionally prefixed to help identify their source, and to avoid starting the string with a leading space, as we noticed that models often drop the leading space after a quotation mark, even for normal tokens. Although many other prompt formats are effective, we have found this code-based approach to more clearly avoid false positives. Figure 5 shows the result for Mistral, Anthropic and OpenAI models.

(a) Mistral API prompting results.

(b) Claude API prompting results.

(c) GPT-3.5 API prompting results.

(d) GPT-4 API prompting results.

Figure 5: API prompting results.